\documentclass[a4paper,fleqn]{cas-dc}

\usepackage[numbers,sort&compress]{natbib}
\usepackage{multirow}
\usepackage{amsmath}
\usepackage{booktabs}
\usepackage{graphicx}

\begin{document}
\let\WriteBookmarks\relax
\def\floatpagepagefraction{1}
\def\textpagefraction{.001}

\shorttitle{Learning Diagnostic Feedback for CGEC}
\shortauthors{Chang et al.}

\title[mode=title]{CriticGEC: Learning Diagnostic Feedback for Chinese Grammatical Error Correction}

\author[1]{Qikai Chang}
\fnmark[1]
\ead{qkchang@mail.ustc.edu.cn}

\author[1]{Pengfei Hu}
\fnmark[1]
\ead{pengfeihu@mail.ustc.edu.cn}

\author[2]{Zhenrong Zhang}
\ead{zrzhang6@iflytek.com}

\author[2]{Jiefeng Ma}
\ead{jfma@iflytek.com}

\author[1]{Jun Du}
\ead{jundu@ustc.edu.cn}

\author[2]{Jianshu Zhang}
\ead{jszhang6@iflytek.com}

\fntext[fn1]{These authors contributed equally.}

\affiliation[1]{organization={University of Science and Technology of China},
  addressline={No. 96 Jinzhai Road},
  city={Hefei},
  state={Anhui},
  country={China}}

\affiliation[2]{organization={iFLYTEK AI Research},
  addressline={No. 666 West Wangjiang Road},
  city={Hefei},
  state={Anhui},
  country={China}}

\begin{abstract}
Chinese grammatical error correction (CGEC) aims to detect and correct errors in Chinese text while preserving the intended meaning. Existing methods primarily produce corrected sentences. However, in educational settings, the goal is not only to provide a correct sentence, but also to diagnose learners' own correction attempts and guide them through further revision. To address this need, we propose CriticGEC, a diagnostic feedback model that examines a candidate correction to identify Undercorrection, Overcorrection and Miscorrection, and generates explanatory natural-language critiques. To train CriticGEC, we automatically construct 253,223 high-quality critique examples from existing CGEC corpora without additional manual annotation. We further introduce two complementary mechanisms for utilizing these critiques. Sequential Revision guides the actor through successive rounds of diagnosis and revision, while Parallel Revision generates multiple critique-guided revisions and retains character-level edits supported by a majority vote. Both mechanisms can be applied separately or combined. Experiments demonstrate improvements across different actors, including Qwen2, Qwen2.5, Qwen3, and a Qwen3-8B actor trained through supervised fine-tuning. With the latter actor and both revision mechanisms, CriticGEC achieves $\text{F}_{0.5}$ scores of 61.69 on FCGEC-test and 63.16 on NaCGEC-test. By providing diagnostic feedback on attempted corrections, CriticGEC offers a foundation for interactive, learner-oriented revision rather than corrected sentences.
\end{abstract}



\maketitle

\section{Introduction}\label{sec:introduction}

Chinese grammatical error correction (CGEC) aims to detect and correct errors in Chinese text while preserving the intended meaning and making as few changes as necessary \cite{bryant2023grammatical}. It has applications in proofreading and writing assistance \cite{ng-etal-2014-conll}. Early approaches relied on handcrafted grammar rules, while subsequent neural methods performed correction by predicting editing operations or directly generating corrected sentences \cite{kwasny-sondheimer-1981-relaxation,omelianchuk-etal-2020-gector,zhao2020maskgec}. More recently, large language models (LLMs) have attracted growing attention in CGEC. Initial work explored zero-shot and few-shot prompting \cite{wu2023chatgpt}. CoTask \cite{liu2025chain} then used multitask instruction tuning to improve the recognition and handling of grammatical errors, EPO \cite{liang2025edit} introduced preference optimization that emphasizes edited tokens, and EARM \cite{earm} combined chain-of-thought reasoning, edit-aware reward modeling, and reinforcement learning to enhance correction performance. Compared with earlier approaches, LLMs can also generate natural-language analyses and, in some cases, explain the reasons for their corrections.

\begin{figure*}[t]
  \centering
  \includegraphics[width=1.8\columnwidth]{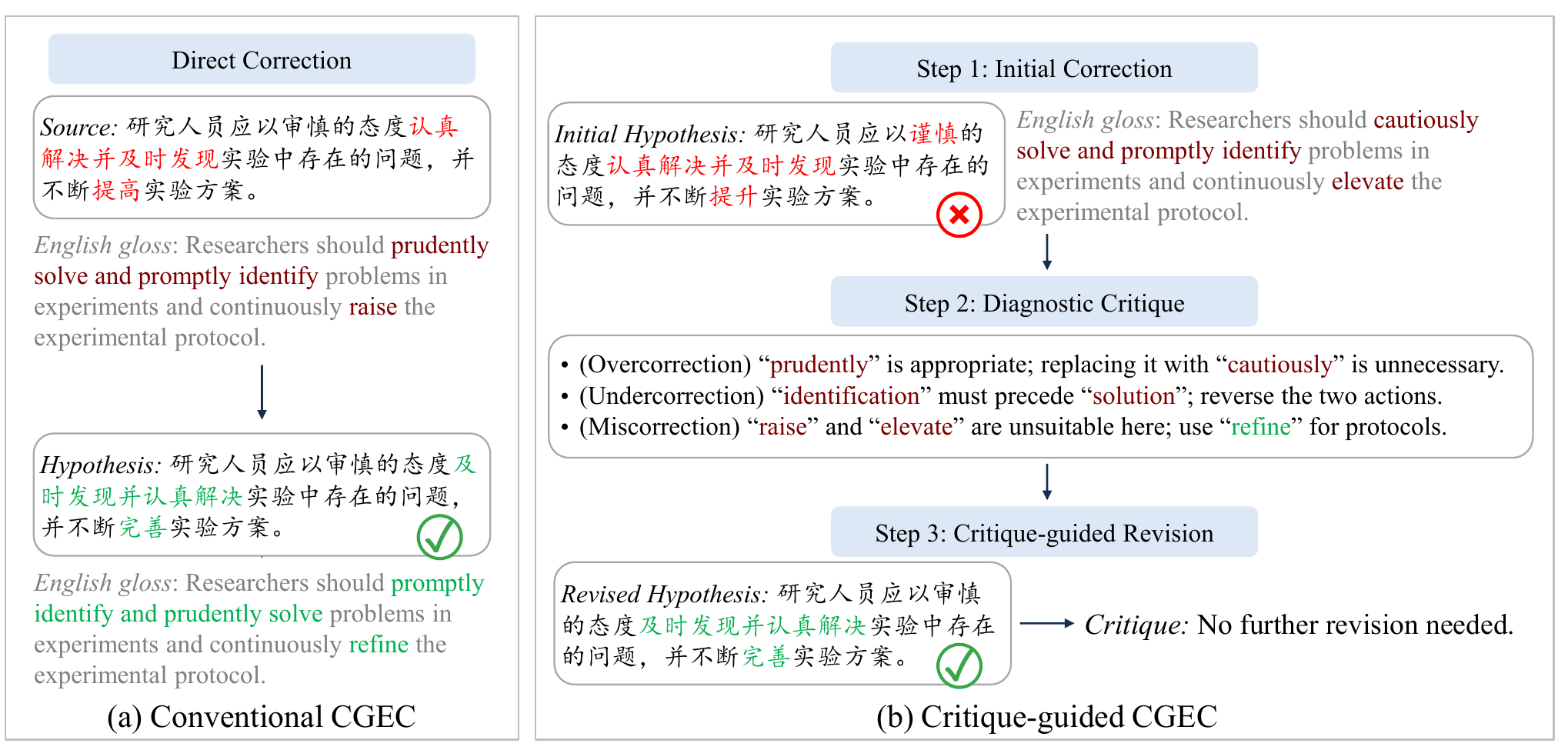}
  \caption{Comparison between conventional CGEC and feedback-guided correction of an attempted revision.}
  \label{fig:gec_teaser}
\end{figure*}

Despite this progress, existing CGEC methods do not fully meet the needs of educational settings that call for diagnosis and guidance. As illustrated in Fig.~\ref{fig:gec_teaser}(a), conventional methods take an original sentence as input and directly produce a corrected version. Some LLM-based methods also explain why particular changes are made. This paradigm is well suited to proofreading and writing assistance when obtaining a corrected sentence is the primary goal. In educational settings, another useful approach is to let learners attempt the correction first and then provide feedback on their own answers \cite{liu2024socraticlm}. Figure~\ref{fig:gec_teaser}(b) illustrates this process. A learner's attempt may be entirely correct, leave an original error unresolved (\emph{Undercorrection}), change text that did not need correction (\emph{Overcorrection}), or identify an error but correct it in an inappropriate way (\emph{Miscorrection}). These problems may also occur together. The system must therefore examine the original sentence alongside the learner's attempt, identify remaining errors, assess whether the learner's changes are necessary and correct, and explain why further revision is needed or why a particular change is inappropriate. Learners can then revise their answers in response to the feedback until no further correction is needed. Rather than simply presenting another corrected sentence, this process emphasizes diagnosing and guiding the learner's own revisions. It also requires the system to recognize not only errors that remain uncorrected but also changes that are unnecessary or incorrect. In fact, 
the use of natural-language feedback to improve initial outputs has been explored in general LLM tasks. For example, Self-Refine \cite{selfrefine} generates feedback on an initial response and uses it to produce a revised one. RL4F \cite{rl4f} trains a feedback model through reinforcement learning to provide critiques that guide a task model. These studies provide a basis for using feedback to guide correction. Applying this approach to the educational setting described above raises three questions: how to identify Undercorrection, Overcorrection, and Miscorrection in a candidate revision by considering the original sentence; how to construct the natural-language critique data needed for training; and how to use the resulting critiques to guide subsequent revisions effectively.

To address these questions, we propose CriticGEC, a method for Chinese grammatical error correction guided by diagnostic natural-language feedback. An existing LLM serves as the \emph{actor} and produces an initial correction, simulating a learner's attempt, while a specially trained critique model, \emph{CriticGEC}, provides the diagnosis. Given the original sentence and the actor's candidate correction, CriticGEC identifies unresolved errors as well as unnecessary or incorrect changes and generates natural-language feedback with suggestions for revision. If the candidate is already correct, it indicates that no further changes are needed. The actor then revises its answer according to the critique. To train CriticGEC, we also develop an automatic data construction process with three stages: collection, annotation, and filtering. This process produces triples consisting of an original sentence, a candidate correction, and its diagnostic critique from existing CGEC corpora, without additional manual critique annotation or fine-grained error-type labels. In total, we construct 253,223 training examples and train a dedicated critique model. In educational applications, the learner could instead take the role of the actor, allowing CriticGEC to provide feedback on the learner's actual correction attempt and support interactive, individualized revision.

To make effective use of these critiques, we further propose two complementary mechanisms: \emph{Sequential Revision} and \emph{Parallel Revision}. In Sequential Revision, the actor and CriticGEC interact over multiple rounds. CriticGEC diagnoses the current candidate, and the actor revises it according to the feedback until the critic accepts the answer or the maximum number of rounds is reached. This process helps the actor resolve issues left unaddressed by an earlier revision and gives the critic further opportunities to identify missed errors. Parallel Revision instead samples multiple critiques independently for the same candidate. The actor generates a revision from each critique, and character-level edits supported by a strict majority are retained to reduce the influence of inconsistent suggestions. The two mechanisms can be used independently or combined.

We conduct experiments on FCGEC \cite{xu2022fcgec} and NaCGEC \cite{ma2022linguistic}. The results show that CriticGEC improves correction performance for different actors, including Qwen2, Qwen2.5, Qwen3, and a Qwen3-8B actor trained through supervised fine-tuning. With both revision mechanisms, the system based on the fine-tuned Qwen3-8B achieves $F_{0.5}$ scores of 61.69 on FCGEC-test and 63.16 on NaCGEC-test, exceeding the corresponding results of CoTask, EPO, and EARM-Single. Case studies further show that CriticGEC can diagnose Undercorrection, Overcorrection, and Miscorrection in an actor's attempts and guide subsequent revisions. These findings demonstrate its potential to provide explanatory and interactive feedback in educational applications.

Our main contributions are summarized as follows:
\begin{itemize}
  \item We introduce CriticGEC, which extends CGEC beyond directly generating a corrected sentence by diagnosing a candidate against its original source, identifying Undercorrection, Overcorrection, and Miscorrection, and providing natural-language suggestions for revision.
  \item We develop an automatic critique data construction process based on collection, annotation, and filtering, yielding 253,223 training examples from existing CGEC corpora without additional manual annotation.
  \item We propose Sequential Revision and Parallel Revision, two complementary mechanisms that use repeated diagnosis and independent critique agreement, respectively, to guide an actor's corrections. They can be applied separately or together and improve performance across different actors.
\end{itemize}

\section{Related Work}\label{sec:related}

\subsection{Edit-based GEC}

Since source and target sentences in grammatical error correction typically share most of their tokens, edit-based methods predict localized editing operations rather than regenerating the entire corrected sentence. Sequence-tagging approaches assign edit labels to input tokens, allowing multiple edits to be applied in parallel. LaserTagger uses a predefined set of token-level operations, including KEEP, DELETE, and APPEND \cite{malmi-etal-2019-encode}. PIE employs parallel iterative editing for local sequence transduction \cite{awasthi-etal-2019-parallel}. GECToR further extends the tag inventory with transformations for common grammatical phenomena \cite{omelianchuk-etal-2020-gector}. Although efficient, these approaches rely on predefined tagging schemes, which may limit their ability to represent open-vocabulary substitutions and complex edits.
To provide more flexible edit representations, sequence-to-edit methods explicitly generate editing operations over spans. Seq2Edits represents edits through span-level operations with explicit position and replacement information, overcoming some limitations of fixed token-level tag inventories \cite{stahlberg-kumar-2020-seq2edits}. Li et al. further formulate GEC as action-guided sequence generation, in which predicted actions are applied to the source sentence \cite{li2022sequence}. Lai et al. explore error-type-driven multi-turn correction, using predicted error types to guide successive editing operations \cite{lai-etal-2022-type}. Despite their efficiency and explicit representation of local changes, edit-based methods remain dependent on the design and coordination of editing operations, particularly when handling complex, context-dependent revisions.

\subsection{Seq2Seq GEC}

Sequence-to-sequence (Seq2Seq) methods formulate GEC as conditional text generation, directly mapping an erroneous sentence to its corrected counterpart. Early neural approaches applied neural machine translation to GEC \cite{yuan-briscoe-2016-grammatical}. Ji et al. subsequently proposed a nested-attention neural hybrid model to improve contextual representation \cite{ji-etal-2017-nested}. Chollampatt and Ng introduced a multilayer convolutional encoder--decoder architecture for grammatical error correction \cite{chollampatt-ng-2018-multilayer}. The Transformer further advanced sequence modeling through self-attention mechanisms \cite{transformer}. Building on this architecture, Zhao et al. introduced a copy-augmented Transformer to exploit the substantial overlap between source and target sentences and preserve unchanged content during decoding \cite{zhao-etal-2019-improving}.
Subsequent research improved neural GEC through synthetic-data pre-training \cite{grundkiewicz-etal-2019-neural} and multilingual transfer learning \cite{rothe-etal-2021-simple}. Meanwhile, pretrained text-to-text models provided more powerful encoder--decoder architectures for generation tasks, including BART \cite{lewis2020bart} and T5 \cite{raffel2020exploring}. For Chinese GEC, SynGEC incorporates dependency syntax to enhance grammatical error correction \cite{zhang2022syngec}. Zhou et al. investigate decoding interventions to improve Seq2Seq correction \cite{zhou2023improving}, while Wang et al. explore multi-turn decoding for Chinese GEC \cite{xiaoying}. Although these methods have substantially advanced automatic correction, they primarily learn a direct mapping from an erroneous sentence to a corrected output. They generally do not explicitly assess an intermediate correction attempt or provide natural-language feedback identifying which edits remain incorrect or unnecessary.

\subsection{LLM-based GEC}

Recent studies have explored large language models (LLMs) for GEC through prompting, instruction tuning, and preference optimization. Evaluations show that ChatGPT can generate fluent corrections but remains prone to overcorrection \cite{fang2023chatgpt,wu2023chatgpt}. To improve task-specific performance, CoTask decomposes Chinese GEC into auxiliary tasks for instruction tuning \cite{liu2025chain}, while Alirector uses source--correction alignment to mitigate overcorrection \cite{alirector}. Other studies explore structured reasoning and iterative self-reflection to improve correction reliability \cite{chang2025chunk,kong2025scholargec}. Preference-based approaches further optimize editing quality: EPO focuses on edit-wise preferences \cite{liang2025edit}, whereas EARM models correction quality at both sentence and token levels \cite{earm}.
Beyond correction generation, recent work has investigated natural-language explanations and interpretable assessment. GEE! generates explanations for individual edits in erroneous--corrected sentence pairs \cite{song-etal-2024-gee}, while Prompt Insertion guides LLMs to explain specific correction points \cite{kaneko-okazaki-2024-prompt-insertion}. For Chinese GEC, EXCGEC provides an explanation-augmented benchmark and investigates pre-explaining and post-explaining settings \cite{ye-etal-2025-excgec}. Other studies explore LLMs as explainers and evaluators \cite{li-etal-2025-rethinking}, use explanations to retrieve relevant correction examples \cite{wang2025re2}, or develop interpretable evaluation metrics and explanation-augmented benchmarks \cite{ye-etal-2025-cleme2,hou2026mecgec}.

Despite these advances, LLM-based GEC remains largely oriented toward producing  corrected sentences. Although recent studies have introduced reasoning, explanation, and assessment capabilities, the correction process is still predominantly treated as a task of obtaining a correct final output. In educational settings, however, effective assistance also requires supporting learners as they revise their own sentences. This involves examining the changes made to the original sentence, identifying both errors that remain and modifications that are unnecessary or inappropriate, and providing targeted feedback to help learners improve their revisions. How to systematically support this feedback-driven revision process remains an important yet underexplored aspect of LLM-based GEC.

\section{Method}\label{sec:method}

\subsection{Training CriticGEC}\label{sec:critic_training}

CriticGEC is designed to identify Undercorrection, Overcorrection, and Miscorrection in an actor prediction and to provide explanatory feedback. Its construction consists of automatic training-data generation followed by supervised fine-tuning.

\subsubsection{Critique Data Construction}

Existing CGEC datasets provide a source sentence and one or more reference corrections. We use the FCGEC training set \cite{xu2022fcgec} and the EXAM subset of NaSGEC \cite{zhang-etal-2023-nasgec}, which contain 35,354 and 4,000 source sentences, respectively. After within-corpus deduplication and removal of samples overlapping the development or test sets, 39,339 source instances remain. Let the base set be $\mathcal{D}=\{(\boldsymbol{s},\mathcal{Y})_i\}_{i=1}^{N_{\rm base}}$, where $\boldsymbol{s}$ is a source sentence and $\mathcal{Y}$ is its set of references. Because these pairs contain no diagnosis of a particular revision, we construct critique supervision using the three-stage pipeline as follows: 


\paragraph{Collection.}
For every source $\boldsymbol{s}$, we repeatedly sample actor outputs, remove duplicates, and obtain $\mathbb{P}=\{\boldsymbol{p}_1,\ldots,\boldsymbol{p}_{J}\}$. To diversify revision behavior, we mix untuned and variants trained through supervised fine-tuning (SFT) of Qwen2, Qwen2.5, and Qwen3. A candidate may be correct, leave a source error unresolved (Undercorrection), change originally correct content (Overcorrection), or repair a detected error in an invalid way (Miscorrection). These categories are not mutually exclusive. A candidate exactly matching any reference is treated as correct. If the source or a reference has not been sampled, it is added to $\mathbb{P}$ to enrich the candidate pool.

\paragraph{Annotation.}
For each candidate $\boldsymbol{p}_j$, we generate a diagnostic critique $\boldsymbol{c}_j$. Exact reference matches receive the predefined termination critique ``The answer is correct.'' All other candidates are annotated by Qwen3-30B-A3B-Thinking-2507 \cite{Yang2025Qwen3TR}. If several references are available, we select the one with the smallest edit distance from $\boldsymbol{p}_j$, denoted $\boldsymbol{y}_j$. Annotation is expressed as
\begin{equation}
\boldsymbol{c}_j=\operatorname{Annotator}(\boldsymbol{s},\boldsymbol{p}_j,\boldsymbol{y}_j).
\end{equation}
The reference serves only as supervision for automatic annotation: it helps the annotator determine which source content requires correction and whether the candidate's edits are justified. The annotator compares the source, candidate, and reference, locates the three types of correction errors, explains their causes, and suggests further revisions. References are never provided to CriticGEC during training or inference; the critic receives only the source and candidate.

\paragraph{Filtering.}
Automatically generated critiques inevitably contain noise. We first remove samples that exceed the maximum decoding length, violate the output format, or leak annotation-only information such as ``reference answer.'' We then test whether a critique can actually guide revision. Given a candidate triple $(\boldsymbol{s},\boldsymbol{p}_j,\boldsymbol{c}_j)$, a Qwen3-8B actor revises the candidate:
\begin{equation}
\boldsymbol{p}'_j=\mathcal{A}(\boldsymbol{s},\boldsymbol{p}_j,\boldsymbol{c}_j).
\end{equation}
We retain a triple only when $\boldsymbol{p}'_j$ exactly matches $\boldsymbol{y}_j$, indicating that the critique effectively guides the validation actor toward the reference correction. To further evaluate the quality of the retained critiques, we conduct a human evaluation, as described in Section~\ref{sec:quality}. The resulting dataset is $\mathcal{D}_{\mathrm{critic}}=\{(\boldsymbol{s},\boldsymbol{p},\boldsymbol{c})_i\}_{i=1}^{N_{\rm critic}}$.

\subsubsection{Dataset Statistics}

All actors generate candidates for the same 39,339 source sentences. Each untuned actor is sampled 16 times per source, whereas each SFT actor produces 16 sampled outputs plus one greedy decoding. To prevent an SFT actor from generating candidates for its own training examples, we use four-fold cross-generation: each actor is trained on three folds and sampled on the held-out fold. Table~\ref{tab:gec_actor_sampling} summarizes the candidate statistics before annotation and filtering. The Candidates column reports the number of distinct candidates after within-source deduplication, while Per source gives their average number per source sentence. The Across actors rows further deduplicate candidates across actors within each setting. The results reveal substantial redundancy in repeated sampling, particularly among newer untuned models. The effect of SFT on candidate diversity varies across model generations, while combining candidates from multiple actors retains complementary outputs and expands the candidate pool. After annotation and filtering, the final dataset contains 253,223 triples, averaging 6.44 triples per source sentence.

\begin{table}[t]
	\centering
	\caption{Candidate statistics before annotation and filtering. \textit{Candidates} denotes the number of distinct candidates after within-source deduplication, and \textit{Per Source} denotes the average number of distinct candidates per source sentence. \textit{Across actors} denotes counts after further deduplication across actors .}
	\label{tab:gec_actor_sampling}
	\resizebox{0.97\columnwidth}{!}{
			\begin{tabular}{lccc}
			\toprule
			Actor & Setting & Candidates  & Per Source\\
			\midrule
			\multirow{2}{*}{Qwen2-7B-Instruct \cite{Yang2024Qwen2TR}}
			& Untuned & 192,368 & 4.89\\
			& SFT & 68,843 & 1.75\\
			\midrule
			\multirow{2}{*}{Qwen2.5-7B-Instruct \cite{Yang2024Qwen25TR}}
			& Untuned & 70,417 & 1.79\\
			& SFT  & 72,777 & 1.85\\
			\midrule
			\multirow{2}{*}{Qwen3-4B \cite{Yang2025Qwen3TR}}
			& Untuned  & 40,126 & 1.02\\
			& SFT  & -- & --\\
			\midrule
			\multirow{2}{*}{Qwen3-8B \cite{Yang2025Qwen3TR}}
			& Untuned  & 45,633 & 1.16\\
			& SFT  & 75,924 & 1.93\\
			\midrule
			\multirow{2}{*}{Across actors}
			& Untuned  & 286,742 & 7.29\\
			& SFT  & 115,617 & 2.94\\
			\bottomrule
			\end{tabular}}
\end{table}

\begin{figure*}[t]
\centering
\includegraphics[width=2\columnwidth]{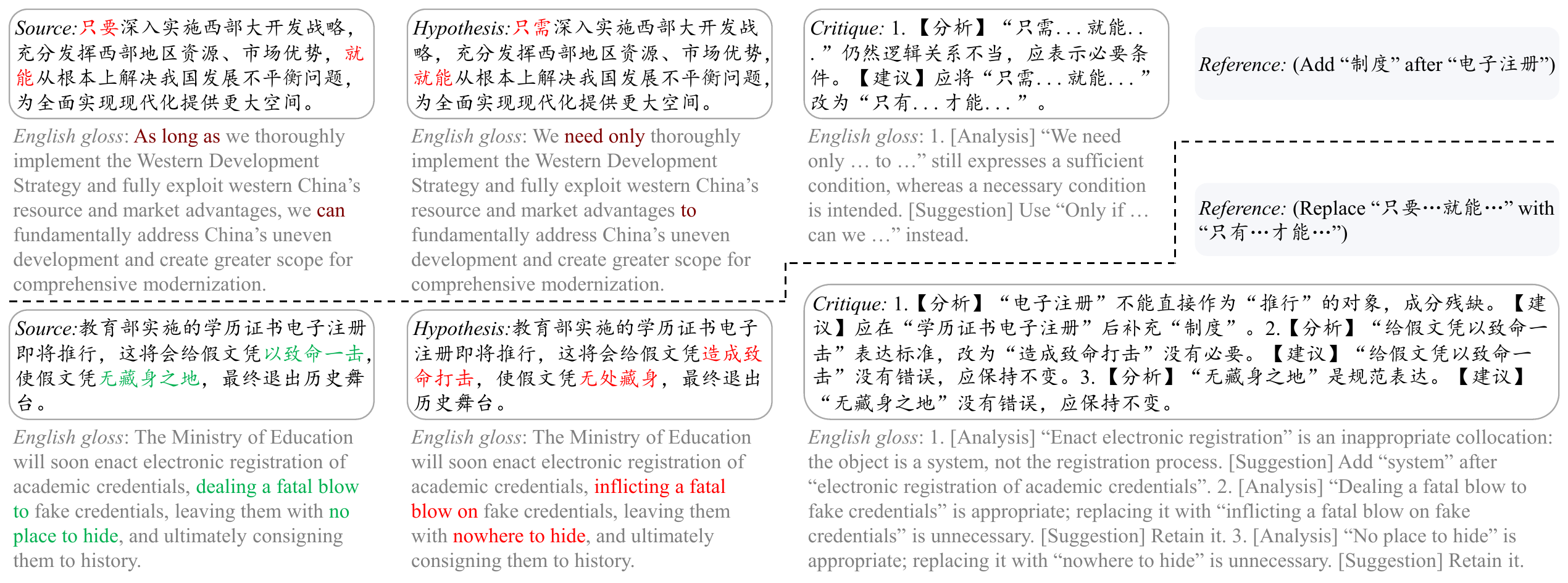}
\caption{Examples of automatically constructed training instances.}
\label{fig:gec_dataset_example}
\end{figure*}

\begin{figure}[pos=t]
\centering
\includegraphics[width=0.9\columnwidth]{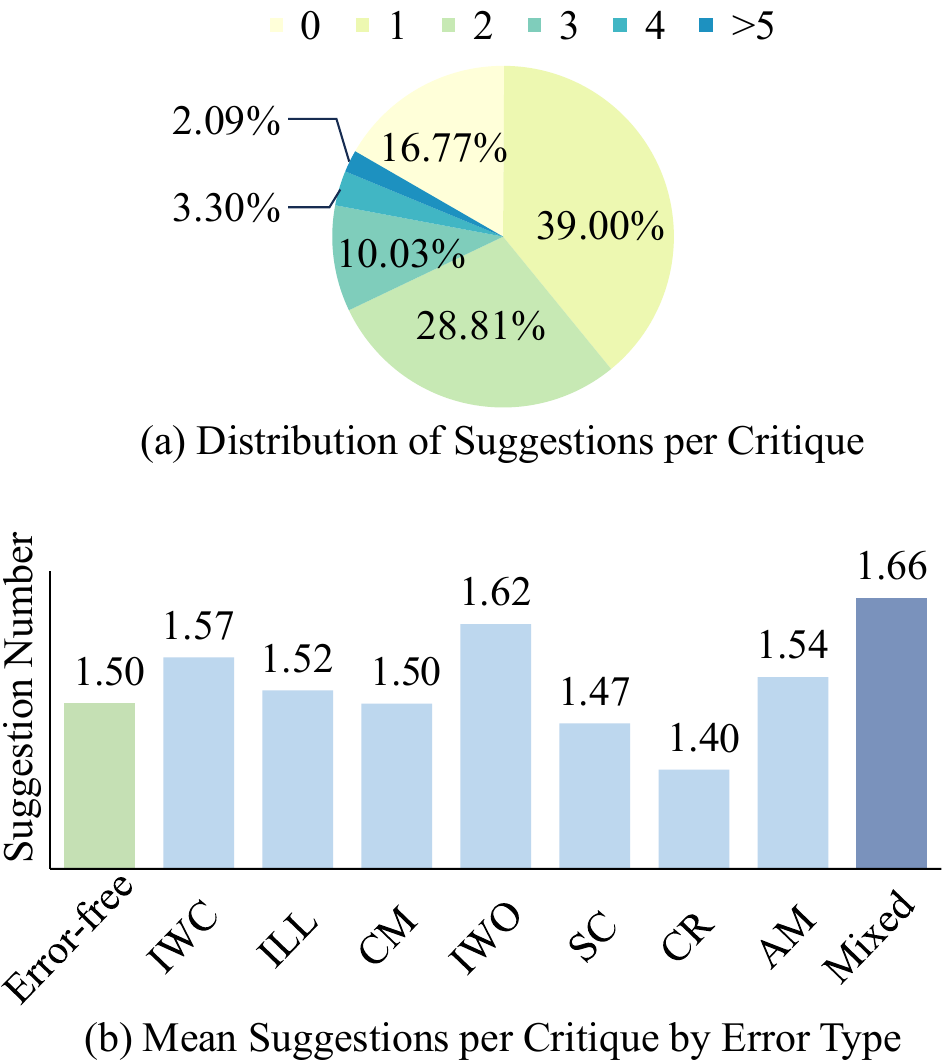}
\caption{Distribution of critique suggestions.}
\label{fig:suggest_num_all}
\end{figure}

Figure~\ref{fig:gec_dataset_example} illustrates two examples of the automatically constructed training instances. The critiques can identify all three types of correction errors: Undercorrection, Overcorrection, and Miscorrection. Beyond identifying problematic edits, they explain why an edit is necessary, inappropriate, or should be retained, and provide corresponding revision suggestions. A single critique can also organize multiple diagnoses and suggestions, explaining the reasons behind different correction decisions. Figure~\ref{fig:suggest_num_all}(a) shows that most critiques contain no more than three suggestions, while zero-suggestion critiques correspond to candidates that are already correct and require no further revision. The small number of critiques containing many suggestions mainly arises from two cases: multiple traditional-to-simplified Chinese character conversions and extensive over-editing of longer sentences. Figure~\ref{fig:suggest_num_all}(b) reports the mean number of suggestions by source error type. IWC, ILL, CM, IWO, SC, CR, and AM denote Incorrect Word Collocation, Illogical, Component Missing, Incorrect Word Order, Structure Confusion, Component Redundancy, and Ambiguity, respectively. Error-free refers to sources without grammatical errors, whereas Mixed refers to sources containing multiple error types. Notably, Error-free sources still receive suggestions on average because actors may introduce unnecessary edits even when the source is correct. Mixed-error sources receive the most suggestions, as they typically involve multiple types of grammatical errors. The critiques average 72.3 Chinese characters, compared with 54.3, 54.5, and 54.7 characters for the source, reference, and candidate sentences, respectively, reflecting the additional space needed to locate errors, explain their causes, and suggest revisions.

\subsubsection{Model Training}

For each instance in $\mathcal{D}_{\mathrm{critic}}$, the source $\boldsymbol{s}$, candidate $\boldsymbol{p}$, and task instruction are concatenated into context $\boldsymbol{x}_{\mathrm{c}}$, and the critique $\boldsymbol{c}$ is the target. For $\boldsymbol{c}=[\boldsymbol{c}_1,\ldots,\boldsymbol{c}_{T_{\mathrm{c}}}]$, CriticGEC is trained autoregressively with cross-entropy:
\begin{equation}
\mathcal{L}=-\frac{1}{T_{\mathrm{c}}}\sum_{t=1}^{T_{\mathrm{c}}}\log P(\boldsymbol{c}_t\mid\boldsymbol{c}_{<t};\boldsymbol{x}_{\mathrm{c}}).
\end{equation}
The resulting CriticGEC can work with different actors.

\subsection{Critique-guided Revision}\label{sec:feedback}

\begin{figure*}[t]
\centering
\includegraphics[width=1.95\columnwidth]{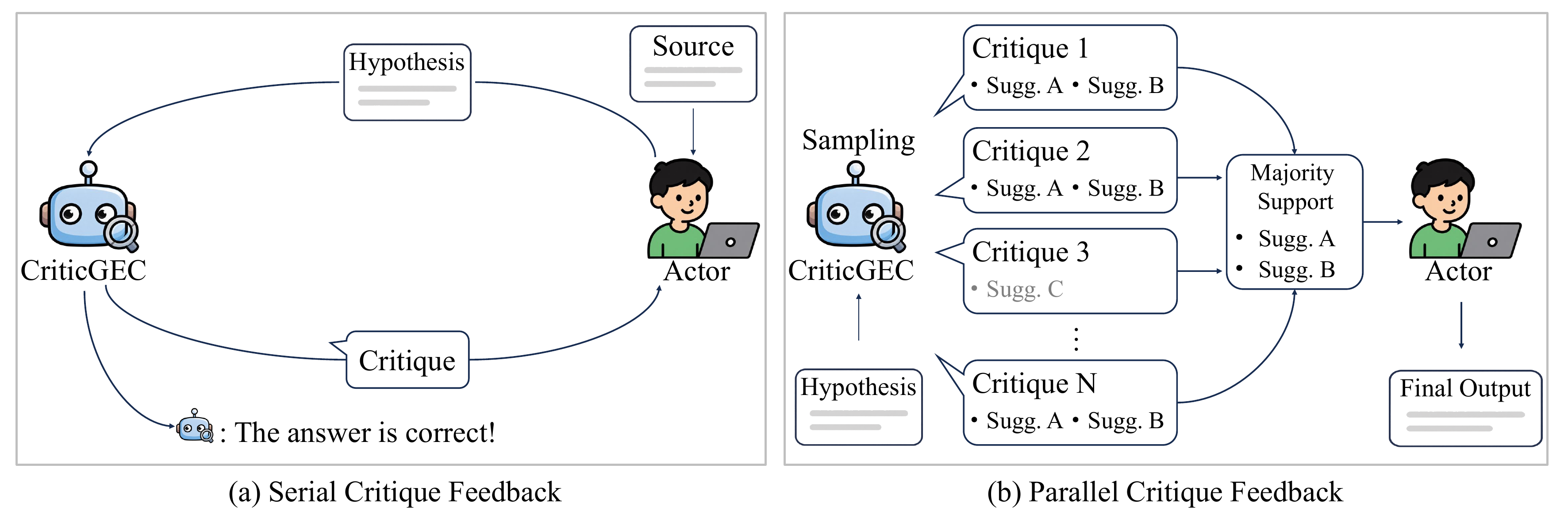}
\caption{Sequential Revision and Parallel Revision using CriticGEC feedback.}
\label{fig:critique_use}
\end{figure*}

Figure~\ref{fig:critique_use} illustrates the two mechanisms. Sequential Revision progressively repairs a prediction through repeated interaction between the actor and CriticGEC. Parallel Revision evaluates the same prediction using independent critiques and retains edits supported by a majority.

\subsubsection{Sequential Revision}

Given a source $\boldsymbol{s}$, the actor first generates
\begin{equation}
\boldsymbol{p}^{0}=\mathcal{A}(\boldsymbol{s}).
\end{equation}
At interaction round $k$, CriticGEC diagnoses the current prediction:
\begin{equation}
\boldsymbol{c}^{k}=\mathcal{C}(\boldsymbol{s},\boldsymbol{p}^{k}).
\end{equation}
If $\boldsymbol{c}^{k}$ is the predefined termination message, the process stops. Otherwise, the actor revises the prediction using the source, current prediction, and critique:
\begin{equation}
\boldsymbol{p}^{k+1}=\mathcal{A}(\boldsymbol{s},\boldsymbol{p}^{k},\boldsymbol{c}^{k}).
\end{equation}
Diagnosis and revision alternate until the critic accepts the prediction or the maximum number of iterations $K_{\max}$ is reached. Later rounds can expose an error missed by a previous critique or repair an actor's incomplete execution of earlier advice. This process also mirrors educational interaction in which feedback is updated after each learner attempt.

\subsubsection{Parallel Revision}

A single sampled critique may overlook an error or recommend an unnecessary edit. Parallel Revision reduces this uncertainty by independently sampling $N$ critiques for the same source and prediction:
\begin{equation}
\boldsymbol{c}_{i}\sim\mathcal{C}(\boldsymbol{s},\boldsymbol{p}),\qquad 1\leq i\leq N.
\end{equation}
Natural-language critiques may express the same advice differently, so direct textual agreement is unreliable. We instead ask the actor to realize each critique as a revised sentence,
\begin{equation}
\boldsymbol{r}_{i}=\mathcal{A}(\boldsymbol{s},\boldsymbol{p},\boldsymbol{c}_{i}),\qquad 1\leq i\leq N,
\end{equation}
and map it to character-level replacements, deletions, and insertions:
\begin{equation}
\mathcal{E}_{i}=\operatorname{Edit}(\boldsymbol{p},\boldsymbol{r}_{i}).
\end{equation}
Edits agree only when their operation type, position, and content are identical. For any $op\in\bigcup_{i=1}^{N}\mathcal{E}_{i}$, its support is
\begin{equation}
n(op)=\sum_{i=1}^{N}I\!\left(op\in\mathcal{E}_{i}\right).
\end{equation}
We retain the strict-majority set
\begin{equation}
\mathcal{V}=\left\{op\in\bigcup_{i=1}^{N}\mathcal{E}_{i}\mid n(op)>\frac{N}{2}\right\}.
\end{equation}
Applying $\mathcal{V}$ to $\boldsymbol{p}$ yields the final revision $\boldsymbol{r}^{*}$. Thus, unstable edits supported by only a minority of critiques are filtered out, whereas repeatedly recommended edits are retained. The two mechanisms address complementary sources of error and may be combined.

\section{Experiments}\label{sec:experiments}

\subsection{Experimental Setup}\label{sec:setup}



CriticGEC is initialized from Qwen3-8B \cite{Yang2025Qwen3TR}, while Qwen3-30B-A3B-Thinking-2507 \cite{Yang2025Qwen3TR} serves as the annotator in thinking mode. CriticGEC is trained with Swift \cite{swift} for three epochs using a learning rate of $3\times10^{-5}$ and a global batch size of 256. We use mixed-precision distributed training and DeepSpeed ZeRO-1 \cite{rasley2020deepspeed} on eight NVIDIA H200 Tensor Core GPUs. The SFT Qwen3 actor is first trained for three epochs on HSK \cite{hsk,zhao2018overview} and Lang-8 \cite{mizumoto2011mining}, with a learning rate of $1\times10^{-5}$ and a global batch size of 256. It is then fine-tuned for one epoch on FCGEC \cite{xu2022fcgec} and the EXAM subset of NaSGEC \cite{zhang-etal-2023-nasgec}, using a learning rate of $2\times10^{-5}$ and the same batch size. Unless stated otherwise, Qwen2, Qwen2.5, and Qwen3 denote Qwen2-7B-Instruct \cite{Yang2024Qwen2TR}, Qwen2.5-7B-Instruct \cite{Yang2024Qwen25TR}, and Qwen3-8B \cite{Yang2025Qwen3TR}; thinking mode is disabled for all LLMs except the annotator.

We use the following inference settings for the LLMs. For CriticGEC, greedy decoding is used in Sequential Revision, while Parallel Revision uses sampling with a temperature of $0.6$, top-$p$ of $0.95$, and top-$k$ of $20$. For the actor, the SFT Qwen3 model uses greedy decoding, whereas untuned actors follow their officially recommended sampling settings. Unless otherwise specified, Sequential Revision terminates early when the critic accepts the answer and uses $K_{\max}$=4; Parallel Revision uses $N$=4. When both mechanisms are combined, we run $N$ independent Sequential Revision trajectories from the same initial actor output and aggregate their final revisions using the majority rule.
We evaluate on FCGEC \cite{xu2022fcgec} and NaCGEC \cite{ma2022linguistic}. The FCGEC-test reference labels are not publicly accessible, so we submitted predictions to the official FCGEC online evaluation system and obtained the reported test-set scores there. All ablation studies and hyperparameter analyses use FCGEC-dev. Following standard CGEC practice, we report precision, recall, and $F_{0.5}$ using ChERRANT \cite{zhang-etal-2022-mucgec}. No post-processing is applied.

\subsection{Human Evaluation of Critiques}\label{sec:quality}

To assess the quality of the automatically constructed critiques, we conducted an independent human evaluation. We randomly sampled 200 triplets from the filtered training data. Two independent annotators with expertise in Chinese grammatical error correction, both native Chinese-speaking graduate students, independently evaluated each triplet along two aspects: (1) \emph{judgment accuracy}, assessing whether the critique correctly judged whether the candidate revision was grammatically correct; and (2) \emph{suggestion validity}, assessing whether the diagnostic analysis and proposed revision advice were linguistically correct and reasonable. In Table~\ref{tab:human_eval}, \emph{Judgment} and \emph{Suggestion} refer to these two evaluation aspects, respectively. \emph{Accuracy} denotes the mean correctness rate across the two annotators, while \emph{Agreement} represents the proportion of instances for which both annotators gave the same judgment. The results suggest that the automatically constructed critiques are generally reliable, providing independent evidence of the quality of the filtered training data.

\begin{table}[t]
	\centering
	\caption{Human evaluation of automatically generated critiques.}
	\label{tab:human_eval}
	\begin{tabular}{lcc}
		\toprule
		Criterion & Accuracy (\%) & Agreement (\%) \\
		\midrule
		Judgment & 99.5 & 99.0\\
		Suggestion & 95.0 & 98.0\\
		\bottomrule
	\end{tabular}
\end{table}

\subsection{Ablation Studies}\label{sec:ablation}

\subsubsection{Effect of Critic Feedback}

\begin{table}[t]
	\centering
	\caption{Effect of different critics on Qwen3-8B.}
	\label{tab:gec_exp_critic}
		\begin{tabular}{lccc}
			\toprule
			\multirow{2}{*}{Critic} & \multicolumn{3}{c}{FCGEC-dev}\\
			\cmidrule(lr){2-4}
			& P & R & $F_{0.5}$\\
			\midrule
			None & 22.73 & 11.53 & 19.03\\
			Qwen3-8B \cite{Yang2025Qwen3TR} & 20.37 & 8.42 & 15.87\\
			$\mathrm{CriticGEC}_{\mathrm{bool}}$ & 20.05 & 14.12 & 18.50\\
			\textbf{CriticGEC} & \textbf{47.62} & \textbf{42.22} & \textbf{46.44}\\
			\bottomrule
		\end{tabular}
\end{table}

We first isolate the effect of the critic with Qwen3 as the actor, $K_{\max}$=1, and without Parallel Revision. We compare direct actor inference, an untuned Qwen3-8B critic, a weakened $\mathrm{CriticGEC}_{\mathrm{bool}}$ that exposes only the binary messages ``Incorrect; please try again'' or ``The answer is correct,'' and the full natural-language CriticGEC. Table~\ref{tab:gec_exp_critic} reports the results.
Full critiques produce large and balanced gains. On FCGEC-dev, $F_{0.5}$ increases from 19.03 to 46.44. Recall on FCGEC-dev more than triples, showing that the detailed guidance helps the actor recover errors missed in its initial output. In contrast, using an untuned Qwen3 as the critic lowers performance, indicating that a generic LLM is not automatically aligned with the evaluation criteria of CGEC. Binary feedback slightly improves recall by forcing reconsideration but sharply reduces precision because it provides neither an error location nor a reason. The task-specific natural-language critique is therefore central to the improvement.



\subsubsection{Generalization Across Actors}

\begin{table}[t]
	\centering
	\caption{Effect of CriticGEC on different actors.}
	\label{tab:gec_exp_actor}
	\resizebox{\columnwidth}{!}{
		\begin{tabular}{llccc}
			\toprule
			\multirow{2}{*}{Actor} & \multirow{2}{*}{CriticGEC} & \multicolumn{3}{c}{FCGEC-dev}\\
			\cmidrule(lr){3-5}
			& & P & R & $F_{0.5}$\\
			\midrule
			Qwen2-7B-Instruct \cite{Yang2024Qwen2TR} & & 3.94 & 17.62 & 4.66\\
			Qwen2-7B-Instruct & \checkmark & 19.80 & 30.31 & 21.27\\
			\midrule
			Qwen2.5-7B-Instruct \cite{Yang2024Qwen25TR} & & 9.62 & 14.60 & 10.32\\
			Qwen2.5-7B-Instruct & \checkmark & 39.10 & 38.80 & 39.04\\
			\midrule
			Qwen3-8B \cite{Yang2025Qwen3TR} & & 22.73 & 11.53 & 19.03\\
			Qwen3-8B & \checkmark & 47.62 & 42.22 & 46.44\\
			\midrule
			SFT Qwen3-8B \cite{Yang2025Qwen3TR} & & \textbf{57.11} & 39.63 & 52.48\\
			SFT Qwen3-8B & \checkmark & 57.03 & \textbf{46.14} & \textbf{54.46}\\
			\bottomrule
		\end{tabular}
	}
\end{table}

Table~\ref{tab:gec_exp_actor} evaluates one feedback round with actors of different generations and capabilities. CriticGEC improves $F_{0.5}$ for every actor. For untuned actors, both precision and recall increase, which means the feedback both recovers missed errors and suppresses unnecessary edits. Stronger actors achieve better final performance because they can interpret and execute complex critiques more faithfully.
The SFT actor is trained on corpora from the same sources as CriticGEC, yet critique still improves its recall substantially with only a marginal precision decrease. This finding suggests that revision-specific diagnosis provides information complementary to direct instruction tuning, rather than merely compensating for a lack of task-specific actor training.

\subsubsection{Sequential Revision}

\begin{figure*}[t]
	\centering
	\includegraphics[width=2\columnwidth]{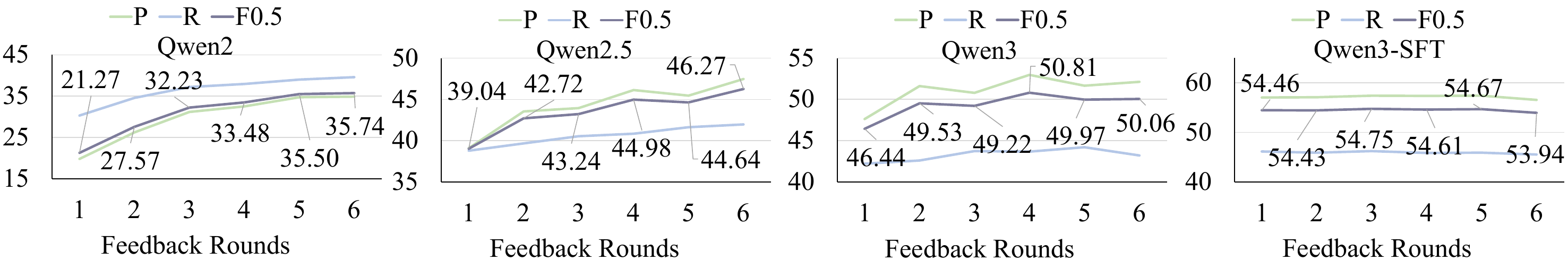}
	\caption{Performance versus the maximum number of feedback rounds.}
	\label{fig:qwens_iter}
\end{figure*}

We vary $K_{\max}$ for Qwen2, Qwen2.5, Qwen3, and SFT Qwen3. The results are plotted in Fig.~\ref{fig:qwens_iter}.
For untuned actors, additional rounds improve correction quality mainly during the first few interactions, after which the gains saturate. One critique can resolve many obvious issues, but later diagnoses can still expose omissions or repair an actor's incomplete response to previous advice. The benefit is larger for weaker actors. For example, Qwen2.5 improves on FCGEC-dev from 39.04 after the first round to 46.27 after six rounds, a gain of 7.23 points. Qwen3 approaches its ceiling in fewer rounds. The SFT Qwen3 actor gains about two points from the first critique but only fluctuates thereafter, indicating that repeated feedback is less useful once the actor is already highly capable.
Early LLMs such as Qwen2 exhibit recall much higher than precision, consistent with prior observations that LLM-based GEC tends to over-correct \cite{wu2023chatgpt}. For Qwen2.5 and Qwen3, precision remains above recall throughout the iterations. Improved instruction following appears to reduce Overcorrection, leaving Undercorrection as a more prominent limitation of modern LLM actors.

\begin{figure}[pos=t]
\centering
\includegraphics[width=1.0\columnwidth]{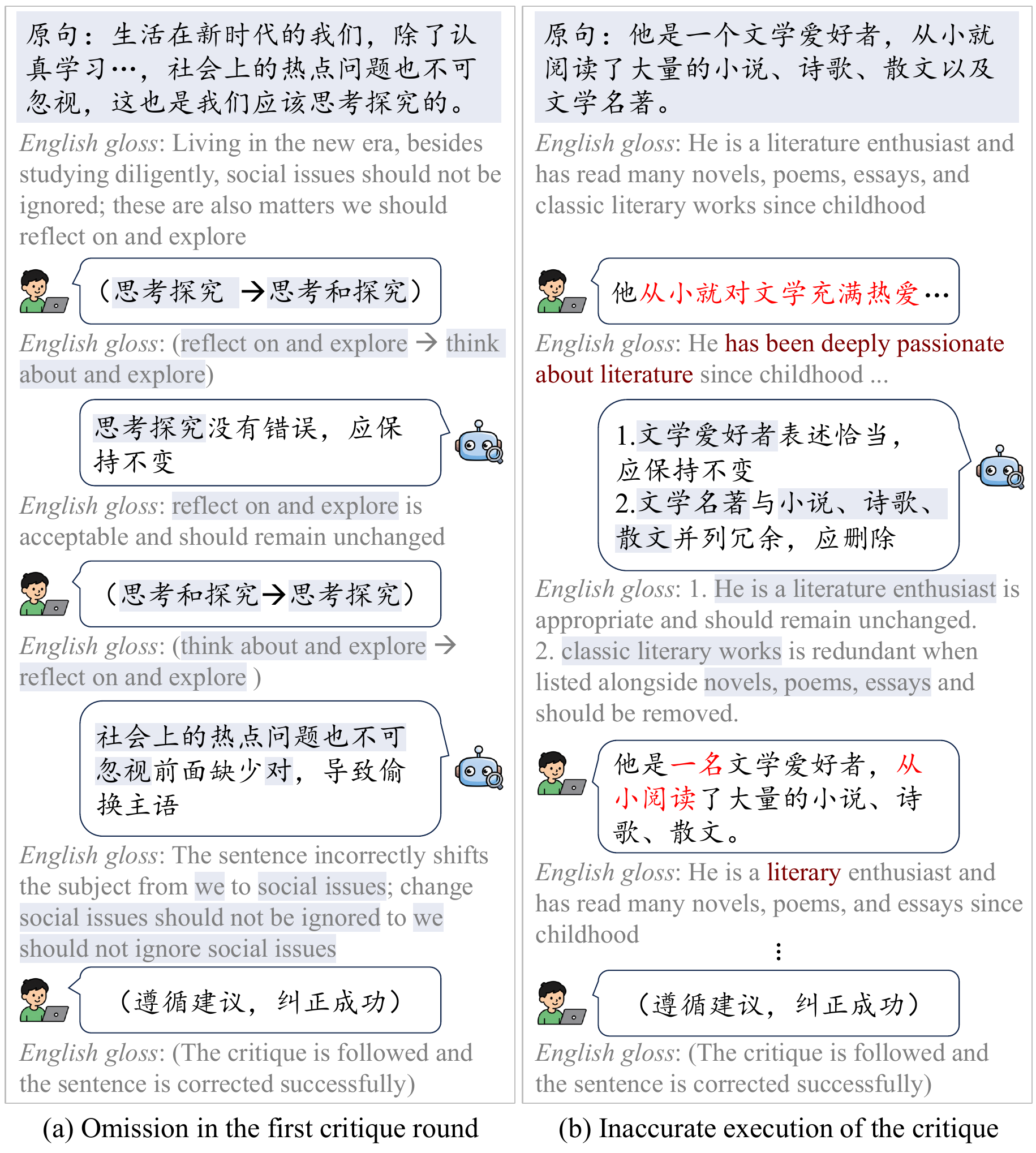}
\caption{Examples of Sequential Revision.}
\label{fig:gec_iter_example}
\end{figure}

Figure~\ref{fig:gec_iter_example} illustrates two roles of Sequential Revision. In the first example, Qwen3 introduces an unnecessary edit and misses a source error. The first critique reverses the Overcorrection, and the second diagnosis then identifies the unresolved missing component. In the second example, Qwen2 incompletely follows a critique and continues introducing extraneous changes, requiring four rounds of redirection. Repeated diagnosis therefore helps both when the critic initially misses an error and when the actor follows correct advice incompletely.

\subsubsection{Parallel Revision}

\begin{figure}[pos=t]
\centering
\includegraphics[width=\columnwidth]{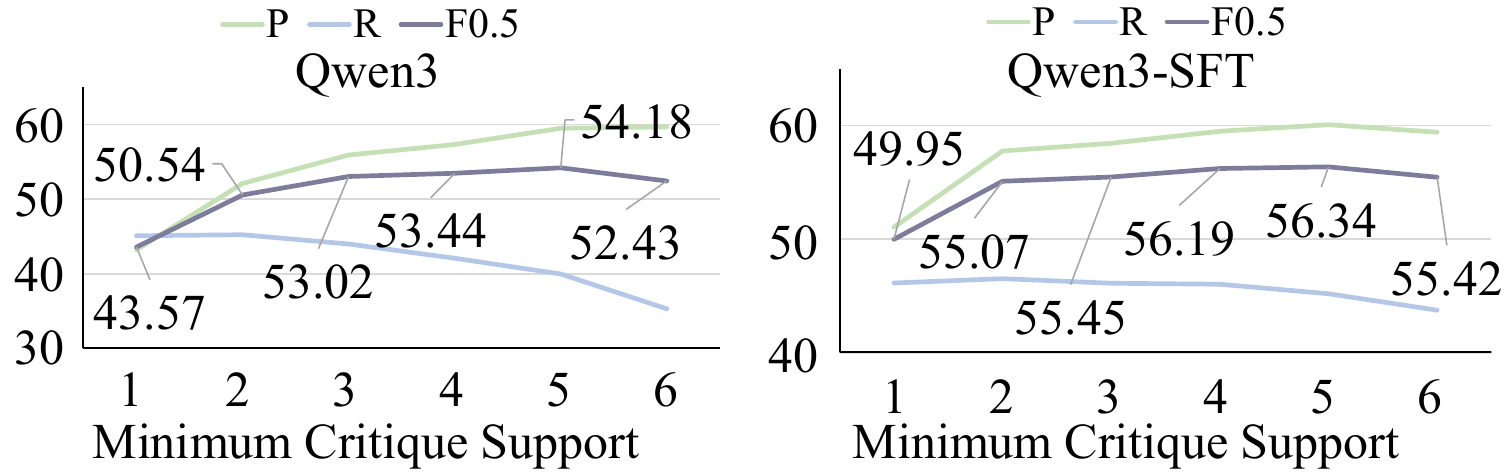}
\caption{Performance versus minimum critique support.}
\label{fig:qwens_min_support}
\end{figure}

Parallel Revision depends on both the minimum number of critiques required to support an edit and the total number of sampled critiques. Let $M$ denote the minimum support and $N$ the number of critiques. We first fix $N$=6 and vary $M$ for Qwen3 and SFT Qwen3. As Fig.~\ref{fig:qwens_min_support} shows, increasing $M$ generally raises precision and lowers recall. A small threshold admits more potentially correct edits but is vulnerable to unstable suggestions; a large threshold is conservative and may discard valid corrections. The best region is near $N/2$, motivating the strict-majority default $M$=$\lfloor N/2\rfloor$+1.

We next vary $N$ from 1 to 6 and set $M$=$\lfloor N/2\rfloor$+1. Table~\ref{tab:gec_exp_num_vote} shows that multiple critiques outperform a single critique by filtering idiosyncratic suggestions. The gains are not monotonic, however: once sufficient independent evidence is available, further samples provide little new information and can occasionally reduce performance. We therefore use $N$=4 as a balance between accuracy and inference cost.

\begin{table}[t]
	\centering
	\caption{Effect of the number of parallel critiques.}
	\label{tab:gec_exp_num_vote}
	\resizebox{0.9\columnwidth}{!}{
	\begin{tabular}{c|c|ccc}
		\toprule
		\multirow{2}{*}{Actor} & \multirow{2}{*}{   \shortstack{Critique\\Count}  } & \multicolumn{3}{c}{FCGEC-dev}\\
		\cmidrule(lr){3-5}
		& & P & R & $F_{0.5}$\\
		\midrule
		\multirow{6}{*}{Qwen3-8B \cite{Yang2025Qwen3TR}}
		& 1 & 47.62 & 42.22 & 46.44\\
		& 2 & \textbf{55.20} & 36.68 & 50.14\\
		& 3 & 53.44 & 41.75 & 50.60\\
		& 4 & 54.78 & 39.98 & 51.01\\
		& 5 & 53.78 & \textbf{42.53} & \textbf{51.08}\\
		& 6 & 52.67 & 40.77 & 49.77\\
		\midrule
		\multirow{6}{*}{  \shortstack{SFT\\Qwen3-8B \cite{Yang2025Qwen3TR}}  }
		& 1 & 57.03 & 46.14 & 54.46\\
		& 2 & 59.01 & 44.77 & 55.48\\
		& 3 & 58.49 & \textbf{46.29} & 55.57\\
		& 4 & \textbf{59.66} & 45.53 & 56.17\\
		& 5 & 58.97 & 46.22 & 55.88\\
		& 6 & 59.49 & 45.99 & \textbf{56.19}\\
		\bottomrule
	\end{tabular}
	}
\end{table}

\subsubsection{Sequential \& Parallel Revision}

\begin{table*}[t]
	\centering
	\caption{Ablation of Sequential Revision and Parallel Revision on FCGEC-dev, with average generated-token counts.}
	\label{tab:gec_exp_feedback}
		\begin{tabular}{l|cc|ccc|c}
			\toprule
			\multirow{2}{*}{Actor} & \multirow{2}{*}{Sequential} & \multirow{2}{*}{Parallel} & \multicolumn{3}{c|}{FCGEC-dev} & \multirow{2}{*}{Avg. tokens}\\
			\cmidrule(lr){4-6}
			& & & P & R & $F_{0.5}$ & \\
			\midrule
			\multirow{4}{*}{Qwen3-8B \cite{Yang2025Qwen3TR}}
			& & & 22.73 & 11.53 & 19.03 & 45.12\\
			& \checkmark & & 52.97 & \textbf{43.68} & 50.81 & 143.92\\
			& & \checkmark & 54.78 & 39.98 & 51.01 & 294.06\\
			& \checkmark & \checkmark & \textbf{57.41} & 42.48 & \textbf{53.64} & 445.81\\
			\midrule
			\multirow{4}{*}{SFT Qwen3-8B \cite{Yang2025Qwen3TR}}
			& & & 57.11 & 39.63 & 52.48 & 44.70\\
			& \checkmark & & 57.36 & \textbf{45.84} & 54.61 & 77.42\\
			& & \checkmark & 59.66 & 45.53 & 56.17 & 152.53\\
			& \checkmark & \checkmark & \textbf{60.95} & 45.07 & \textbf{56.94} & 180.64\\
			\bottomrule
		\end{tabular}
\end{table*}

Table~\ref{tab:gec_exp_feedback} compares the mechanisms separately and jointly using their default hyperparameters: $K_{\max}$=4, $N$=4, and $M$=3. The average generated-token count includes all actor and CriticGEC generations required for one input.
Both mechanisms improve the untuned and SFT actors, but with different profiles. Sequential Revision tends to increase recall by discovering problems omitted in earlier rounds, whereas Parallel Revision favors precision by retaining edits with independent support. On FCGEC-dev, their combination achieves the highest $F_{0.5}$ in both actor settings: 53.64 for Qwen3 and 56.94 for SFT Qwen3. These development-set results motivate using the combined configuration for the final test-set comparison in Section~\ref{sec:comparison}.

\begin{figure}[pos=t]
	\centering
	\includegraphics[width=\columnwidth]{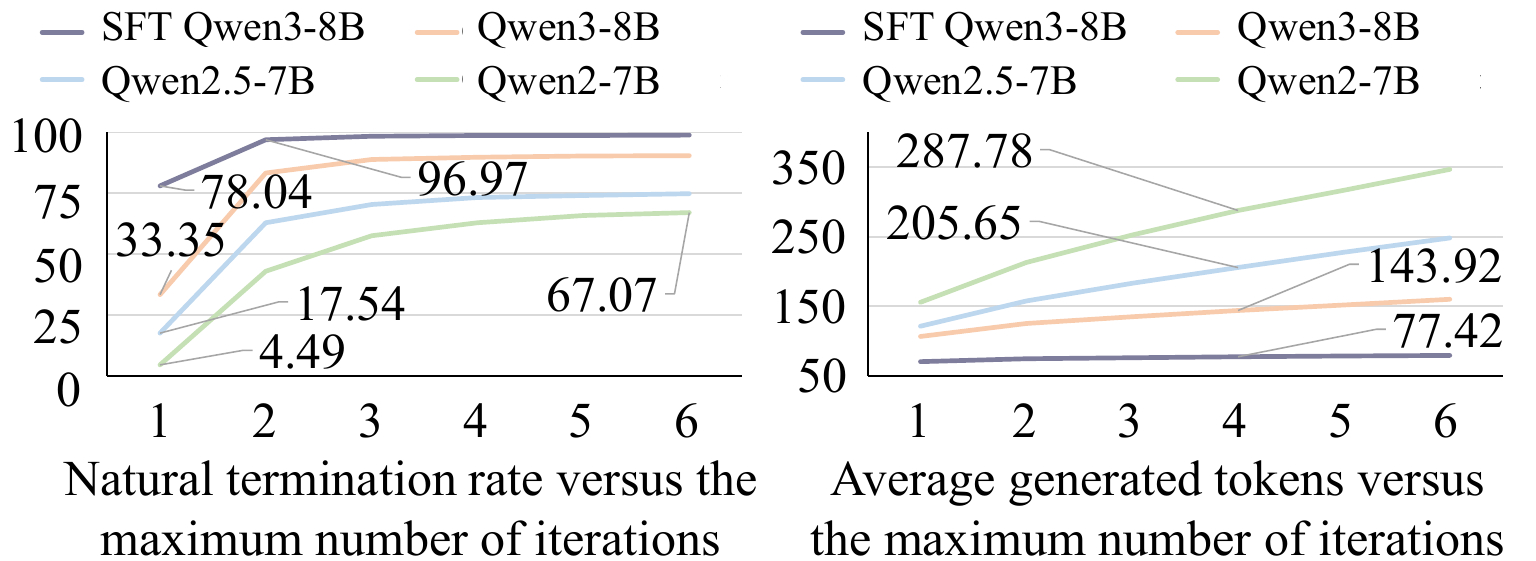}
	\caption{Natural termination rate and average generated tokens in Sequential Revision.}
	\label{fig:gec_serial_cost}
\end{figure}

This improvement introduces additional generation. Direct Qwen3 and SFT Qwen3 produce an average of 45.12 and 44.70 tokens, whereas Sequential Revision increases these values to 143.92 and 77.42. The lower cost for the stronger SFT actor results from early termination: CriticGEC accepts a larger proportion of its initial or early-round answers. Figure~\ref{fig:gec_serial_cost}(a) shows the natural termination rate as $K_{\max}$ increases, where natural termination means that the critic emits the predefined acceptance message. At $K_{\max}=1$, 78.04\% of SFT Qwen3 predictions are accepted, compared with only 4.49\% for Qwen2. By round two, 96.97\% of SFT Qwen3 instances have terminated, whereas more than 30\% of Qwen2 instances remain unaccepted even after six rounds. Correspondingly, Fig.~\ref{fig:gec_serial_cost}(b) shows that token usage grows rapidly in early rounds and then slows rather than increasing linearly with $K_{\max}$.
The termination rate does not reach 100\%. Besides failures to follow advice, some examples reveal a critic-side convergence problem. In one sentence about core technologies and national security, CriticGEC correctly detects a missing component in its first round, but subsequent critiques alternate between two contextually acceptable completions equivalent to ``a foundation for'' and ``a safeguard for.'' Both yield grammatically plausible corrections, yet the critic repeatedly proposes replacing one with the other instead of accepting the revision. This oscillation suggests that critiques can be inconsistent when acceptable expressions are semantically close, potentially exacerbated by the limited coverage of reference corrections in the training data. Tracking previous revisions could prevent such cycles in practice, but we do not apply a rule-based stopping heuristic here so as to examine the model's intrinsic termination behavior.

\subsection{Comparison with Existing GEC Methods}\label{sec:comparison}

\begin{table*}[htp]
	\centering
	\caption{Comparison on FCGEC-test and NaCGEC-test.}
	\label{tab:gec_exp_main}
	\begin{tabular}{l|ccc|ccc}
		\toprule
		\multirow{2}{*}{Model} & \multicolumn{3}{c|}{FCGEC-test} & \multicolumn{3}{c}{NaCGEC-test}\\
		\cmidrule(lr){2-4}\cmidrule(lr){5-7}
		& P & R & $F_{0.5}$ & P & R & $F_{0.5}$\\
		\midrule
		\multicolumn{7}{c}{\textit{Traditional task-specific models}}\\
		GECToR \cite{omelianchuk-etal-2020-gector} & 46.11 & 34.35 & 43.16 & -- & -- & --\\
		BART \cite{lewis2020bart} & 63.07 & 39.95 & 56.53 & 62.04 & 45.84 & 57.94\\
		SynGEC \cite{zhang2022syngec} & 63.75 & 39.78 & 56.89 & 62.42 & 47.41 & 58.71\\
		MrGEC \cite{liu2024towards} & 65.71 & 37.78 & 57.22 & -- & -- & --\\
		MTD \cite{xiaoying} & -- & -- & -- & 68.82 & 44.64 & 62.09\\
		\midrule
		\multicolumn{7}{c}{\textit{Closed-source LLMs}}\\
		GPT-4o-mini \cite{gpt4o} & 18.39 & 14.81 & 17.54 & 11.75 & 11.42 & 11.68\\
		GPT-4o \cite{gpt4o} & 22.33 & 24.14 & 22.67 & 20.21 & 22.78 & 20.67\\
		\midrule
		\multicolumn{7}{c}{\textit{Open-source LLM-based models}}\\
		DeepSeek-R1 \cite{guo2025deepseek} & 18.78 & 36.06 & 20.77 & 19.31 & 37.30 & 21.37\\
		Chunk-CoT \cite{chang2025chunk} & 46.21 & 36.94 & 44.00 & 63.46 & 44.78 & 58.57\\
		Alirector \cite{alirector} & 65.76 & 39.52 & 58.05 & 66.93 & 46.59 & 61.55\\
		CoTask \cite{liu2025chain} & 65.65 & 36.49 & 56.60 & 62.50 & 40.72 & 56.46\\
		EPO \cite{liang2025edit} & 66.67 & 41.93 & 59.63 & 67.09 & 49.97 & 62.79\\
		EARM-Single \cite{earm} & \textbf{68.98} & 42.39 & 61.29 & \textbf{69.78} & 45.57 & 63.08\\
		\midrule
		\textbf{CriticGEC} & 67.33 & \textbf{46.22} & \textbf{61.69} & 66.77 & \textbf{51.92} & \textbf{63.16}\\
		\bottomrule
	\end{tabular}
\end{table*}

We compare CriticGEC with three groups of methods in  Table \ref{tab:gec_exp_main}, emphasizing correction quality alongside its revision-oriented feedback. Traditional task-specific systems include GECToR \cite{omelianchuk-etal-2020-gector}, BART \cite{lewis2020bart}, SynGEC \cite{zhang2022syngec}, MTD \cite{xiaoying}, and the multi-reference MrGEC \cite{liu2024towards}. Closed-source LLM baselines include GPT-4o-mini and GPT-4o \cite{gpt4o}. Open-source LLM-based methods include Chunk-CoT \cite{chang2025chunk}, which uses explicit analysis and chunked reasoning; Alirector \cite{alirector}, which uses alignments between the source and an initial correction; CoTask \cite{liu2025chain}, which applies multi-task instruction tuning; and the reinforcement-learning methods EPO \cite{liang2025edit} and EARM \cite{earm}. The Alirector results are reproduced from EPO \cite{liang2025edit}; DeepSeek-R1 \cite{guo2025deepseek} results are from EARM \cite{earm}; and the GPT-4o-mini and GPT-4o results are from Chunk-CoT \cite{chang2025chunk} and CoTask \cite{liu2025chain}, respectively.
CriticGEC uses SFT Qwen3-8B with the combination of Sequential Revision and Parallel Revision, $K_{\max}$=4, and $N$=4. Unlike the FCGEC-dev results used to select and analyze these mechanisms, the final comparison reports performance on FCGEC-test and NaCGEC-test. The FCGEC-test scores were obtained by submitting our predictions to the official online evaluation system. A dash denotes an unreported result.
CriticGEC achieves $F_{0.5}$ scores of 61.69 and 63.16, exceeding EPO by 2.06 and 0.37 points and EARM-Single by 0.40 and 0.08 points on the two test sets. EARM-Single has higher precision, whereas CriticGEC has substantially higher recall. Beyond final-answer accuracy, CriticGEC generates a diagnosis tied to the actor's current attempt and can use it for continued revision, a capability relevant to interactive educational assessment. 

The strongest LLM methods use different forms of test-time computation. CriticGEC spends computation on critiques and revisions; its combined setting averages 180.64 generated tokens, 4.04 times the 44.70 tokens of direct SFT inference. EPO does not report token counts but uses beam search with width 10, maintaining and expanding multiple hypotheses. EARM first generates an explicit grammatical-analysis chain of thought and then the corrected sentence. These costs are not directly comparable, but CriticGEC's overhead is of a similar practical order. EARM also reports a best-of-16 setting that samples and reranks 16 outputs, reaching 65.04 and 64.59 $F_{0.5}$ on the two datasets. Because this configuration incurs substantially greater candidate-generation and reranking cost, Table~\ref{tab:gec_exp_main} uses EARM-Single as the primary comparison.

\section{Conclusion}\label{sec:conclusion}

We presented CriticGEC, which diagnoses an attempted Chinese grammatical correction by considering the original sentence and candidate together. The model distinguishes Undercorrection, Overcorrection, and Miscorrection and provides natural-language suggestions for further revision. An automatic process constructs 253,223 training examples without additional manual annotation. Sequential Revision uses successive rounds of diagnosis to address unresolved errors, while Parallel Revision selects edits supported by a majority of independent critiques. The two mechanisms can be applied separately or combined.
Experiments on FCGEC and NaCGEC show that CriticGEC improves different Qwen actors and achieves competitive test-set correction accuracy. More importantly, it provides feedback on a specific attempt rather than only returning a corrected answer. This opens up potential educational applications, such as identifying issues in learners' writing and providing explanations and suggestions through iterative interactions. Future work will evaluate the educational value of such interactions with learners and improve the critic's ability to accommodate multiple equally valid corrections.

\section*{Declaration of generative AI and AI-assisted technologies in the manuscript preparation process}
During the preparation of this manuscript, the authors used ChatGPT to assist with language editing. The authors are responsible for reviewing and editing AI-assisted material and take full responsibility for the content of the publication.

\bibliographystyle{cas-model2-names}
\bibliography{ref}

\end{document}